\documentclass[]{spie}  

\usepackage{amsmath,amsfonts,amssymb}
\usepackage[colorlinks=true, allcolors=blue]{hyperref}
\usepackage{subcaption}
\usepackage{graphicx, floatrow}

\title {Observer variation-aware medical image segmentation by combining deep learning and surrogate-assisted genetic algorithms}

\author[a]{Arkadiy Dushatskiy}
\author[b]{Adri\"enne M. Mendrik}
\author[a,c]{Peter A.N. Bosman}
\author[d]{Tanja Alderliesten}

\affil[a]{Life Sciences \& Health group, Centrum Wiskunde \& Informatica, Science Park 123, 1098 XG, Amsterdam, the Netherlands}
\affil[b]{Netherlands eScience Center, Science Park 140, 1098 XG, Amsterdam, the Netherlands}
\affil[c]{Delft University of Technology, Van Mourik Broekmanweg 6, 2628 XE, Delft, the Netherlands}
\affil[d]{Department of radiation oncology, Amsterdam UMC, University of Amsterdam, Meibergdreef 9, 1105 AZ, Amsterdam, the Netherlands}

\authorinfo{Further author information: (Send correspondence to Arkadiy Dushatskiy)\\Arkadiy Dushatskiy: E-mail: arkadiy.dushatskiy@cwi.nl, Telephone: +31 20 592 4073 \newline
T. Alderliesten  is  currently  working  at  the  Department  of  Radiation  Oncology,  Leiden  University  Medical  Center, Albinusdreef 2, 2333 ZA, Leiden, the Netherlands.}

\begin{document} 
\maketitle

\renewcommand{\labelenumii}{\theenumii}
\renewcommand{\theenumii}{\theenumi.\arabic{enumii}.}

\begin{abstract}
There has recently been great progress in automatic segmentation of medical images with deep learning algorithms. In most works observer variation is acknowledged to be a problem as it makes training data heterogeneous but so far no attempts have been made to explicitly capture this variation. Here, we propose an approach capable of mimicking different styles of segmentation, which potentially can improve quality and clinical acceptance of automatic segmentation methods.
In this work, instead of training one neural network on all available data, we train several neural networks on subgroups of data belonging to different segmentation variations separately. Because a priori it may be unclear what styles of segmentation exist in the data and because different styles do not necessarily map one-on-one to different observers, the subgroups should be automatically determined. We achieve this by searching for the best data partition with a genetic algorithm. Therefore, each network can learn a specific style of segmentation from grouped training data. We provide proof of principle results for open-sourced prostate segmentation MRI data with simulated observer variations. Our approach provides an improvement of up to 23\% (depending on simulated variations) in terms of Dice and surface Dice coefficients compared to one network trained on all data.
\end{abstract}


\section{INTRODUCTION}

\subsection{Background}
Applying deep learning algorithms to automatically segment medical images is widely studied due to the significant advances and great success of deep learning algorithms in other fields. Fundamental improvement in segmentation performance was achieved by introducing encoder-decoder models, starting with the U-Net architecture \cite{unet}. The main idea of neural networks of this type is using an encoder to obtain the compressed image features and then to upsample the compressed representation by a decoder. In all such architectures the encoder contains multiple convolutional layers, while the decoder consists of either transposed convolutional or bilinear upsampling layers. These networks can be used for both binary and multi-class segmentation problems. The architecture of the encoder-decoder model has many modification possibilities, varying in number and types of layers \cite{linknet, fcnet, enet}. Deep learning algorithms are widely applied to MRI and CT segmentation. The most commonly considered regions are brain, lungs and lower abdomen, though there are works on upper abdomen and heart segmentation as well \cite{dl_review}. Despite good results of deep learning algorithms in terms of typical metrics, there is an issue of clinical acceptance of automatic segmentations to be tackled. For instance, it was shown, that a convolutional neural network can provide a high-quality automatic delineation of the prostate, rectum and bladder with Dice scores of $0.87$, $0.89$ and $0.95$ respectively, but for clinical usage $80\%$ of these automatically made segmentations required manual correction \cite{prostate_dl_results}. 

The common approach is to train one network on all available data without acknowledging the heterogeneous nature of medical data caused by inter- and intra-observer variation \cite{review2}. One of the recently introduced approaches aware of observer variation is the probabilistic U-Net \cite{probabilistic-unet}. However, the probabilistic U-Net does not aim at capturing observer variations. The main idea of the probabilistic U-Net is to use latent variables, which parameterize the space of the produced predictions. The goal of this model is to produce a set of segmentation predictions, covering the space of probable segmentation masks as much as possible. However, such approach is different from producing several variants of segmentations, each corresponding to a particular segmentation style. Results on lung nodule segmentation show that the probabilistic U-Net produces a variety of possible masks, different in size, but it is virtually impossible to match the predictions with segmentation styles of particular observers \cite{probabilistic-unet}. Thus, this approach does not solve the problem of automatically producing several segmentation masks corresponding to different segmentation styles.

\subsection{The proposed approach}
\label{sec:intro}  
This work is focused on the task of automatic segmentation of medical images with deep learning algorithms while capturing observer variation. Though our approach can be extended to other sources of variation in data, e.g., different scanning devices, we focus on observer variation and
, particularly, consider the following scenario: 
\begin{enumerate}
\item In a dataset, there are multiple scans, one scan per patient. 
\item Each scan is segmented by one observer, the name of the observer is unknown. 
\item The observers may have different ways of segmenting organs (different styles of segmentation) as well as slight variations within one style of segmentation (i.e., intra-observer variation).
\end{enumerate}

We propose, instead of training one neural network on all data, to separately train several neural networks on different subsets of data corresponding to different segmentation styles. These subsets are obtained by an optimization procedure. Such an approach has two major merits: (1) We improve the segmentation quality as each network is trained on (more) homogeneous data and therefore it can more accurately learn each segmentation style rather than learning one average style. (2) Multiple variants of segmentation can be presented to a doctor and it is more likely that they agree with one of the variants than with an average segmentation. These two points may well contribute to the ultimate goal of our work - increased acceptance of automatic segmentation in clinical practice.

\section{Method}
\subsection{Algorithm outline}
\begin{figure}[h]
\floatsetup{valign=t, heightadjust=all}
\ffigbox{%
\begin{subfloatrow}
\ffigbox{\includegraphics[scale=0.05]{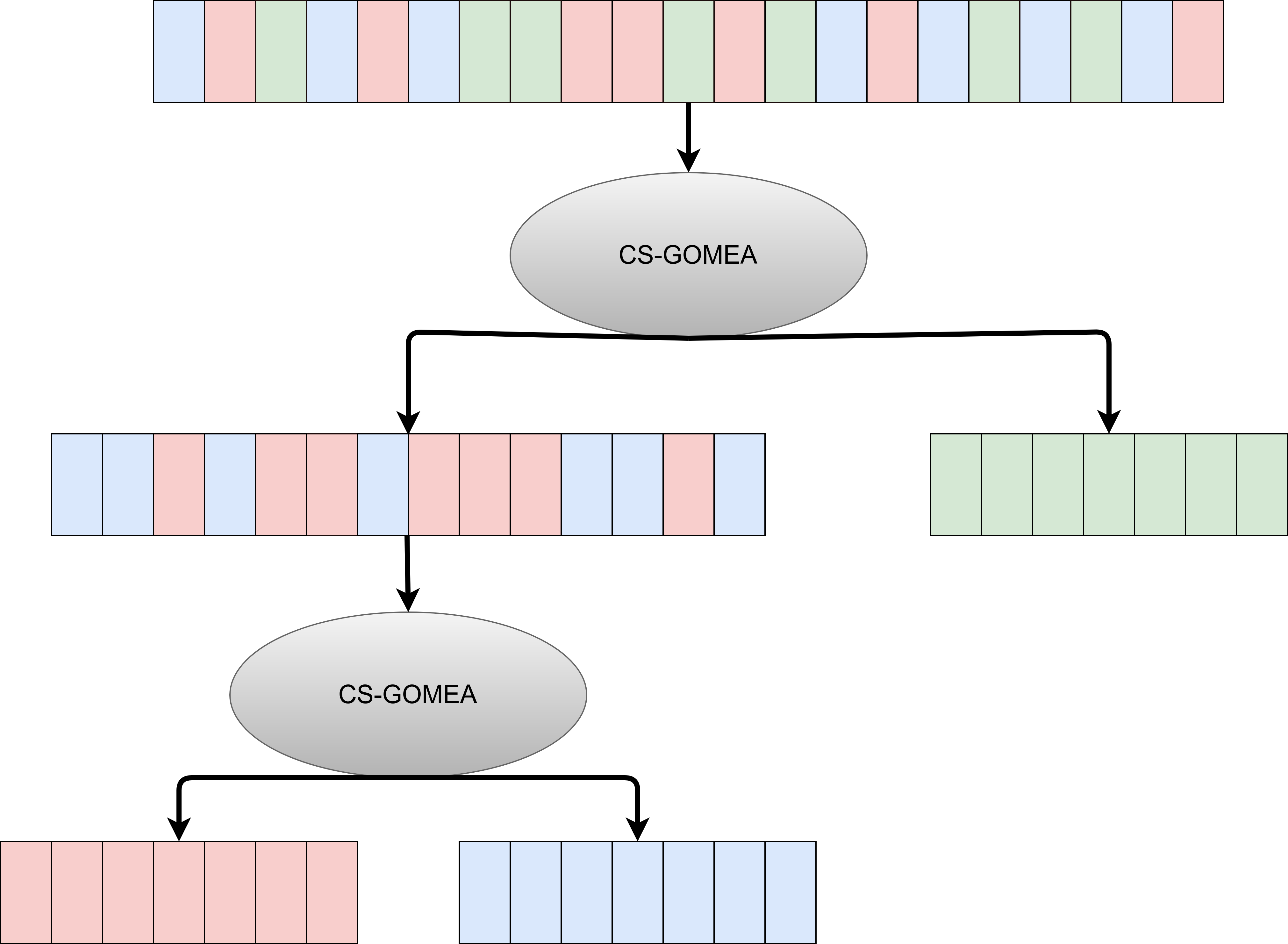}}{\caption{The recursive scheme of the partitioning algorithm. Different colors denote different segmentation styles. The example demonstrates the perfect partition found by a binary optimization algorithm (we use the surrogate-assisted genetic algorithm CS-GOMEA \cite{csgomea}), resulting in three subgroups of scans delineated in three different segmentation styles.} \label{fig:recursion}}
\ffigbox{\includegraphics[scale=0.05]{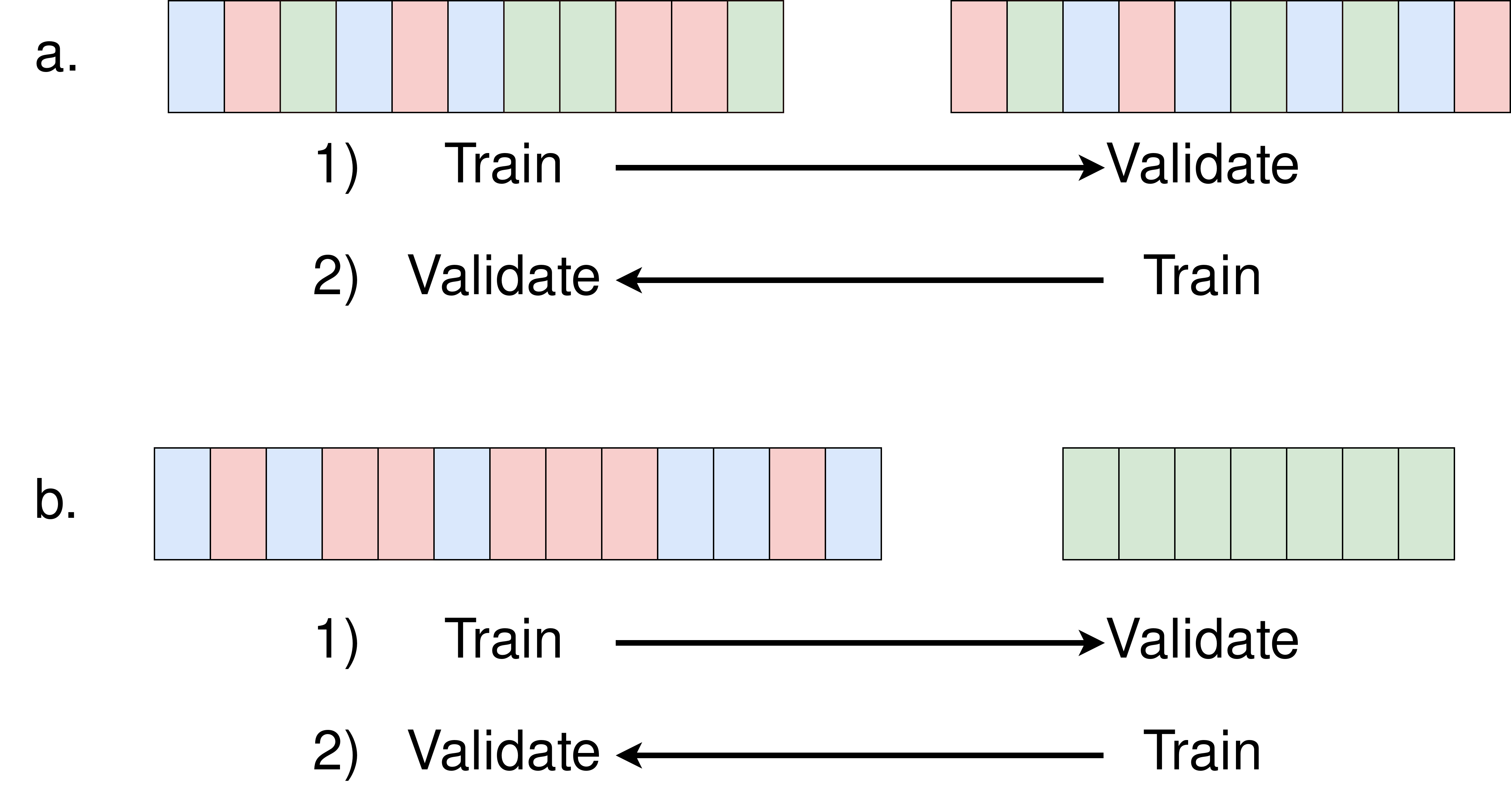}}{\caption{Two examples of a data partitioning evaluated by the optimized objective function. Different colors denote different segmentation styles. The partitioning evaluation contains of two model trainings. First, the scans from the first subgroup are used for training and for the scans from the second subgroup the validation scores are calculated (step 1). Then, the scans from the second subgroup are used for training and the scans from the first subgroup are used as validation set (step 2). The score of the objective function is supposed to be better in the example \textbf{b}, as the subgroups are more different from each other by contained segmentation styles.} \label{fig:objective}}
    \end{subfloatrow}}
    {\caption{The key components of the proposed data partitioning algorithm.}
    \label{fig:algo}}

\vspace{-2cm}

\end{figure}

We assume that there are multiple segmentation styles. Thus, we need to partition $N$ scans in disjoint subgroups, representing different styles of segmentation. We tackle this partitioning task by solving multiple optimization problems (details of the optimization procedure are provided in Section \ref{sub:optim}). To simplify the partitioning task, we propose to solve it hierarchically: at each step divide the scans into two subgroups, and recursively apply the partitioning algorithm. The partitioning algorithm steps are the following:
\begin{enumerate}
    \item Make a partition of scans in two subgroups by solving an optimization problem.
    \item Validate the results: calculate scores of holdout samples in leave-one-out cross-validation on the initial data (a mixture of segmentation styles), and on both found subgroups. If training on the subgroups provides a better score compared to the training on the mixture of data, step 1 is repeated for both subgroups. Otherwise, the partitioning algorithm is stopped.
\end{enumerate}
The scheme of the proposed data partitioning algorithm is provided in Fig.~\ref{fig:recursion}. The algorithm can be terminated when the obtained number of subgroups reaches the assumed number of segmentation styles in a dataset.

\subsection{Segmentation quality evaluation} \label{sub:evaluation}

In the literature on automatic segmentation methods, the Dice-S\o rensen Coefficient (DSC) is a common metric for evaluation: $DSC=\frac{2|G \cap P|}{|G|+|P|}$, where $G$ is the ground truth segmentation mask, $P$ is the predicted mask \cite{dice}.
However, it has drawbacks for automatic segmentation evaluation in clinical practice since (1) DSC values depend on organ size, (2) DSC values do not say anything about the contour part that requires manual correction.
Thus, we also consider the $2D$ version of the surface Dice-Sørensen Coefficient \cite{surfacedice} (SDSC), a recently proposed metric aiming at alleviating shortcomings of the DSC.

The SDSC indicates the percentage of the contour that deviates from the ground truth by more than $\tau$ mm (the margin width $\tau$ is a chosen constant). Formally, $SDSC=\frac{|S_G \cap B_P^\tau|+|S_P \cap B_G^\tau|}{|S_G|+|S_P|}$, where $S_G=\partial G \text{~and~} S_P=\partial P$ are borders of corresponding masks (in the $3D$ case, borders of surfaces), $B_G^\tau, B_P^\tau$ are borders of corresponding masks (surfaces) at a given tolerance $\tau$, $B_G^\tau=\{x\in \Re^2 | \exists \sigma \in S_G,\|x-\xi(\sigma)\| \le \tau\}$,  $B_P^\tau=\{x\in \Re^2 | \exists \sigma \in S_P,\|x-\xi(\sigma)\| \le \tau\}, \xi: S \mapsto \Re^2$.
We use a fixed value of threshold $\tau=0.5mm$ (making it less than voxel size) as it allows to detect small differences between contours. For the optimization procedures, we use the SDSC only, in the final evaluation of the obtained partition we calculate both DSC and SDSC cross-validation scores to demonstrate that there is an improvement in terms of the DSC as well.

\subsection{Optimization procedure} \label{sub:optim}
We formulate the problem of partitioning a set of scans $P$ ($|P|$=N) in two disjoint subgroups $P_1$ and $P_2$, $P_1 \cup P_2 = P$, $P_1 \cap P_2 = \emptyset$ as an optimization problem. The search space (space of solutions) is the space of binary vectors of length $N$. Binary vectors of size $N$ determine the partitioning: a $0$ in position i determines that the i-th scan belongs to $P_1$, a $1$ determines that the i-th scan belongs to $P_2$.

The ultimate goal is to maximize cross-validation scores of two networks trained on a data partition determined by a binary vector. However, the leave-one-out cross-validation approach has a major drawback: it is very computationally expensive (it requires $N$ model trainings) and not scalable since the number of required model trainings is equal to the number of scans. Cross-validation with a fixed number of folds $k$ is less computationally expensive (requires $2k$ model trainings), but our preliminary experiments showed that it is not a reliable method to evaluate a partition as the scores vary a lot across different cross-validation splits. Instead of a direct evaluation of a partition  (i.e., a solution of the formulated optimization problem), we propose to evaluate partitions with a proxy function. The calculation of the straightforward objective {$\boldsymbol{F}$} (normalized cross-validation score) is done as follows:

\begin{enumerate}
    \item \relax[Preliminary step]
    
    Perform leave-one-out cross-validation on $P$ (contains a mixture of styles). For each scan, calculate its hold-out SDSC $M_i$.
\item \relax[$\boldsymbol{F}$ calculation] \begin{enumerate}
    \item Perform leave-one-cross validation on $P_1$; calculate validation SDSC scores $S_i$; obtain relative scores $R_i=\frac{S_i}{M_i}$
    \item Perform leave-one-cross validation on $P_2$; calculate validation SDSC scores $S_i$; obtain relative scores $R_i=\frac{S_i}{M_i}$
    \item The direct function value is an average over collected relative scores $R_i$ (of all scans): $F=\frac{1}{N}\sum_{i=1}^N R_i$
\end{enumerate}
\end{enumerate}

The calculation of the proposed proxy objective {$\boldsymbol{G}$} is done as follows:
\begin{enumerate}
    \item \relax[Preliminary step]
    
    Perform leave-one-out cross-validation on $P$ (contains a mixture of styles). For each scan, calculate its hold-out SDSC $M_i$.
\item  \relax[{$\boldsymbol{G}$} calculation] \begin{enumerate}
\item Train a neural network on $P_1$; calculate validation SDSC scores $S_i, i \in P_2$; obtain relative scores $R_i=\frac{S_i}{M_i}$
    \item Train a neural network on $P_2$; calculate validation SDSC scores $S_i, i \in P_1$; obtain relative scores $R_i=\frac{S_i}{M_i}$
    \item The proxy function value is an average over collected relative scores $R_i$ (of all scans): $G=\frac{1}{N}\sum_{i=1}^N R_i$
\end{enumerate}
\end{enumerate}

The proxy function {$\boldsymbol{G}$} is the objective, which is minimized by a binary optimization algorithm. The scheme of this function calculation is shown in Fig.~\ref{fig:objective}. The intuition behind the proposed proxy function is the following: the more different the segmentation styles of two subgroups, the smaller the scores obtained in steps 2.1 and 2.2 of {$\boldsymbol{G}$} calculation. Thus, we can minimize this proxy function instead of directly maximizing the cross-validation scores. Our experiments show, that it is a valid proxy function, i.e., its values correlate well with the target function. Experiments and results demonstrating this are described in Section \ref{sec:results}. Moreover, the number of neural network trainings is fixed (only two trainings are required), independent of the number of scans.

For optimization, we use a surrogate-assisted genetic algorithm, namely, the recently introduced convolutional neural network surrogate-assisted GOMEA (CS-GOMEA) \cite{csgomea}. This algorithm was chosen as it is capable of finding good solutions within a limited number of function evaluations and it outperforms competitors on a set of benchmark functions. In the current application, the algorithm requires $\approx$ 250 evaluations to converge, which is $\approx$ 5 times less than a state-of-the-art non-surrogate-assisted genetic algorithm (GOMEA).

\subsection{Neural network architecture and training} \label{sub:nnet}
We use the architecture of the Context Encoder Network (CE-Net) \cite{cenet}, which was recently introduced and demonstrated state-of-the-art results on medical image segmentation benchmarks. This architecture falls under the category of encoder-decoder models for segmentation tasks, introduced with the U-Net architecture. The encoder of this network is part of the ResNet-34 architecture, pretrained on the ImageNet dataset. To use the encoder with three input channels without any modifications on grayscale images, we simply copy the input image three times and pass the copies to the encoder.

For neural network training with back-propagation, we use the Adam optimizer \cite{adam}. The learning rate is $10^{-4}$ (lower than the default $10^{-3}$ to make the training more stable). The performance on the validation set is checked every training epoch, and the learning rate is decreased by a factor of 10 if the validation metric has not improved for 5 consecutive epochs. If the learning rate becomes smaller than $10^{-5}$, the training is stopped.
During training, we use random data augmentations to reduce overfitting: scale (by a factor from 0.7 to 1.3), shift (from -40 to 40 px in each direction), rotation (from -10 to 10 degrees), contrast and brightness adjustments (by a factor from 0.5 to 1.5). These augmentations are chosen according to possible scan appearance variations in the dataset.
The batch size is 128. The loss function is the commonly used soft dice loss.

\section{Experiments}
\subsection{Data}
We use open data from the Medical Segmentation Decathlon \cite{decathlon} (\href{http://medicaldecathlon.com}{http://medicaldecathlon.com}) containing MRI scans of the prostate of 32 patients with corresponding segmentation masks. Each segmentation mask represents pixel-wise labels of one of three classes: background, prostate central zone (CZ) and prostate peripheral zone (PZ).
We do not distinguish between the prostate central zone (CZ) and the prostate peripheral zone (PZ) and consider a segmentation task with 2 classes: prostate and background.
To make a proper transformation of the initial $3D$ data to $2D$ slices, we perform a standard procedure of voxel spacing normalization. The resulting voxel spacing of our scans is (0.6mm, 0.6mm, 2mm). After normalization, each scan is zoomed-in by a factor of two to decrease the number of background pixels and rescaled to (128px, 128px) size.

\subsection{Simulated variations}
\begin{figure}[h]
\centering
\begin{subfigure}{.23\textwidth}
    \captionsetup{justification=centering,margin=0.3cm}
    \centering
    \captionsetup{justification=centering,margin=0.3cm}
    \includegraphics[height=3.5cm]{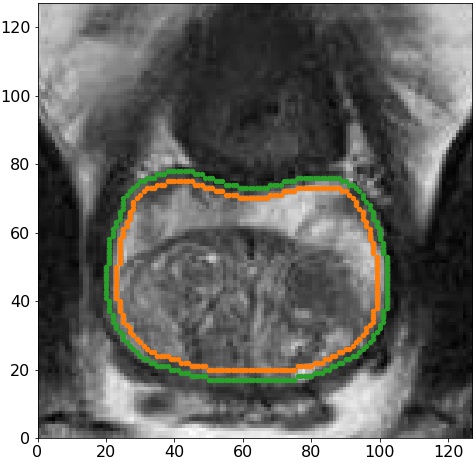}
    \caption[short]{Erosion}
\end{subfigure}%
\begin{subfigure}{.23\textwidth}
    \centering
    \captionsetup{justification=centering,margin=0.3cm}
    \includegraphics[height=3.5cm]{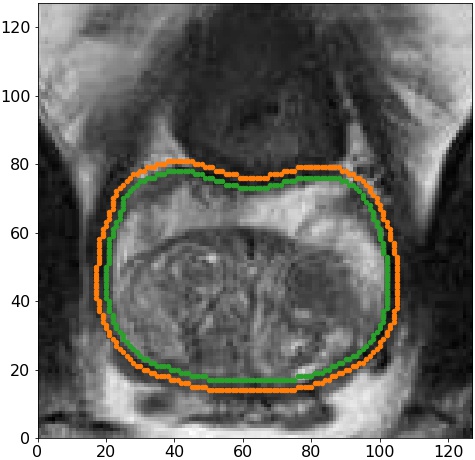}
    \caption[short]{Dilation}
\end{subfigure}
\begin{subfigure}{.23\textwidth}
    \centering
    \captionsetup{justification=centering,margin=0.3cm}
    \includegraphics[height=3.5cm]{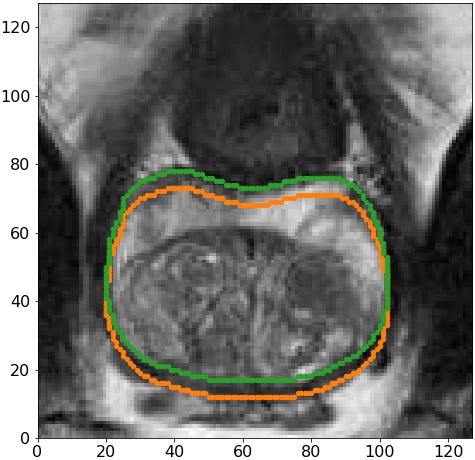}
    \caption[short]{Shift down}
\end{subfigure}
\begin{subfigure}{.23\textwidth}
    \centering
    \captionsetup{justification=centering,margin=0.3cm}
    \includegraphics[height=3.5cm]{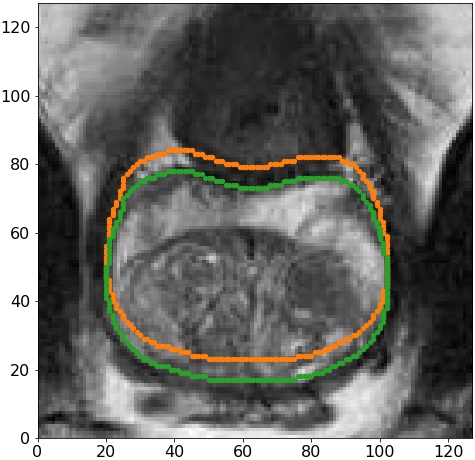}
    \caption[short]{Shift up}
\end{subfigure}%
\vspace{0.5cm}
\begin{subfigure}{.23\textwidth}
    \centering
    \captionsetup{justification=centering,margin=0.3cm}
    \includegraphics[height=3.5cm]{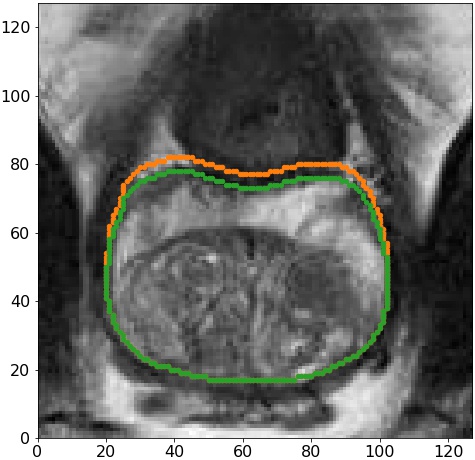}
    \caption[short]{Top \\over-segmentation}
\end{subfigure}
\begin{subfigure}{.23\textwidth}
    \centering
    \captionsetup{justification=centering,margin=0.3cm}
    \includegraphics[height=3.5cm]{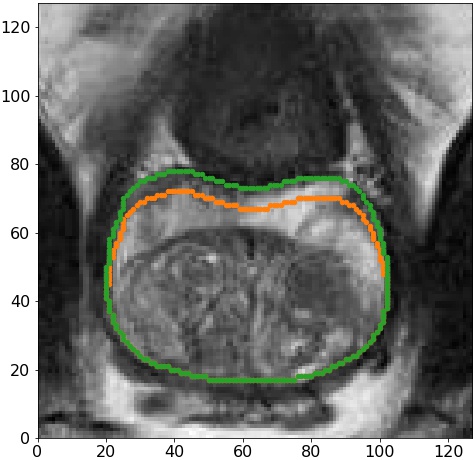}
    \caption[short]{Top \\under-segmentation}
\end{subfigure}
\begin{subfigure}{.23\textwidth}
    \centering
    \captionsetup{justification=centering,margin=0.3cm}
    \includegraphics[height=3.5cm]{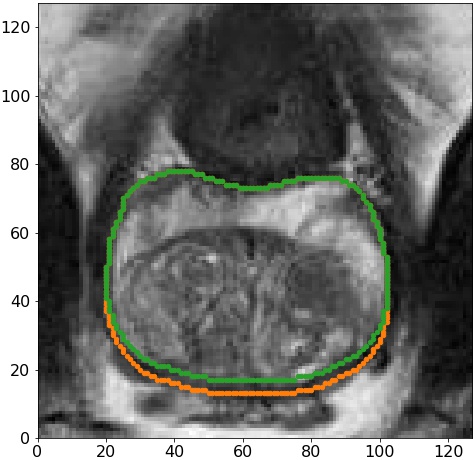}
    \caption[short]{Bottom \\over-segmentation}
\end{subfigure}
\begin{subfigure}{.23\textwidth}
    \captionsetup{justification=centering,margin=0.3cm}
    \includegraphics[height=3.5cm]{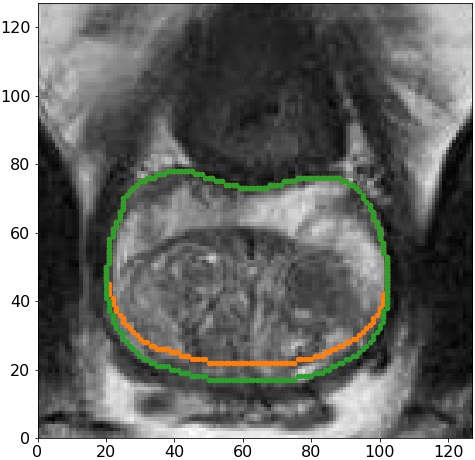}
    \caption[short]{Bottom \\under-segmentation}
\end{subfigure}
\begin{subfigure}{\textwidth}
    \vspace{0.3cm}
    \centering
    \captionsetup{justification=centering,margin=0.3cm}
    \includegraphics[height=1cm]{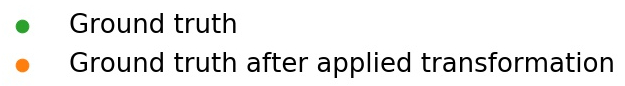}
\end{subfigure}

\caption[short]{Variations applied to the ground truth masks demonstrated on the same slice. The original ground truth contour is plotted in orange, the contour after applying the transformation is illustrated in green. The global variations are shown in the first row, the local ones in the second row. The magnitude of all transformations in these examples was sampled from the Gaussian distribution $\mathcal{N}(5,1)$.}
\label{fig:var}
\end{figure}

We simulated both global and local variations in segmentation styles by applying a transformation to the ground truth prostate segmentation mask. Global variations include erosion, dilation, and shift operations. Local ones are under/over-segmentation of top/bottom (base/apex) parts of the prostate. The chosen variations transform the ground truth contours in prostate zones that have the largest inter-observer variation\cite{prostate_variations}: base and apex.  
Table~\ref{tab:variations} summarizes the applied variations.
Each transformation has a numerical parameter (e.g., shift by $t$ pixels), which is sampled from a normal distribution for each $2D$ slice.
Different transformations simulate different segmentation styles, while the parameters of transformations determine the extent of variations within a style. For instance, we simulate that some scans are segmented by an observer who always over-segments the upper part of the prostate. Other scans are segmented by an observer who under-segments this part (two different styles of segmenting). The magnitude (in pixels) of over- and under-segmentation is sampled from a Gaussian distribution, representing the variation within a style. In Figure~\ref{fig:var} examples of applied variations are illustrated. Transformations are done once, stored on disk and then all experiments are conducted using the obtained data.

\begin{table}[h]
\begin{tabular}{llllllll}
\cline{2-3}
\multicolumn{1}{l|}{}  & \multicolumn{1}{l|}{Global variations} & \multicolumn{1}{l|}{Local variations} &  &  &  &  &  \\ \cline{1-3}
\multicolumn{1}{|l|}{Prostate mask size is changed} & \multicolumn{1}{l|}{erosion} & \multicolumn{1}{l|}{over-segmentation of top part} &  &  &  &  &  \\
\multicolumn{1}{|l|}{} & \multicolumn{1}{l|}{dilation} & \multicolumn{1}{l|}{under-segmentation of top part} &  &  &  &  &  \\
\multicolumn{1}{|l|}{} & \multicolumn{1}{l|}{} & \multicolumn{1}{l|}{over-segmentation of bottom part} &  &  &  &  &  \\
\multicolumn{1}{|l|}{} & \multicolumn{1}{l|}{} & \multicolumn{1}{l|}{under-segmentation of bottom part} &  &  &  &  &  \\ \cline{1-3}
\multicolumn{1}{|l|}{Prostate mask size is unchanged} & \multicolumn{1}{l|}{shift} & \multicolumn{1}{l|}{-} &  &  &  &  &  \\
\multicolumn{1}{|l|}{} & \multicolumn{1}{l|}{} & \multicolumn{1}{l|}{} &  &  &  &  &  \\ \cline{1-3}
                       &                       &                       &  &  &  &  & 
\end{tabular}
\caption{Overview of applied artificial variations.}
\label{tab:variations}
\end{table}


\begin{figure}[h]
    \centering
    \includegraphics[width=\textwidth]{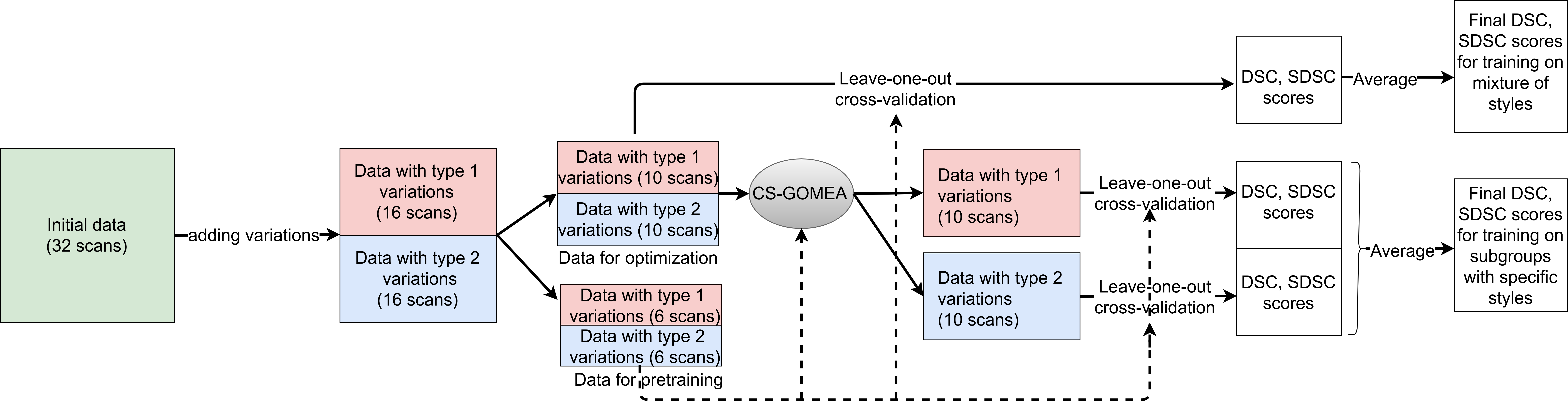}
    \caption{Experiment scheme for the case of two simulated variations. First, two styles of segmentation are simulated for two subgroups of scans. Second, a part of the scans is put aside for pretraining. All networks are pretrained on these scans (indicated by dashed arrows). Then, CS-GOMEA is used to find an optimal partition of the remaining scans. As a result, it produces two subgroups of scans, ideally each having scans with one segmentation style only. To check the improvement provided by the obtained partition, leave-one-out cross-validation is performed on both subgroups and final DSC and SDSC scores are averaged over all scans. They are compared with the DSC and SDSC scores of the leave-one-out cross-validation of training on the mixture of styles. The results of comparison are provided in Table \ref{tab:results}.}
    \label{fig:dataflow}
\end{figure}

To conduct an experiment with two types of variations, we divide all scans in two subgroups (each containing 16 scans). The first subgroup contains scans with applied variation of the first type only, the second subgroup contains scans with variation of the second type. After forming the subgroups with different variations, all scans are gathered together and form a mixture of segmentation styles. Then, we put aside a part of all scans (12 scans) for network pretraining. Our preliminary experiments have shown that pretraining is essential, as it makes the training procedure more stable and helps in solving the optimization task. The remaining scans (20) are used in the optimization algorithm. The scheme of experiments with two segmentation styles is demonstrated in Fig.~\ref{fig:dataflow}.

To demonstrate that our approach is not limited to only two styles, we conduct an experiment with three variations in the data as well. In the case of three segmentation styles, we use 11 scans for pretraining and solve a partitioning problem for 21 scans. In this case there are three subgroups (each has seven scans), each containing scans with a particular type of variation.

We make experiments with eight different combinations of two variations and one combination of three variations. The tested combinations of variations are listed in Table \ref{tab:results}.

\subsection{Objective functions analysis} \label{subsec:obj_analysis}
To investigate how the proxy objective function $\boldsymbol{G}$ described in \ref{sub:optim} is connected with the cross-validation scores function $\boldsymbol{F}$ (direct objective), we do the following experiment. We select two mixtures of scans: one with large (erosion/dilation $\sim\mathcal{N}(10,4)$) and one with small (top over-/under-segmentation $\sim\mathcal{N}(5,1)$) artificial variations. We take the optimal solution $S$ that represents the perfect partition in two subgroups: each subgroup contains scans with variations of one type only. Then, we generate solutions having Hamming distance to $S$ from $1$ to $10$, generating five random solutions for each value of the Hamming distance. For $S$ and these generated solutions we calculate values of $\boldsymbol{F}$ and $\boldsymbol{G}$. 

\section{Results} \label{sec:results}
\begin{figure}[h]
\includegraphics[width=0.49\textwidth]{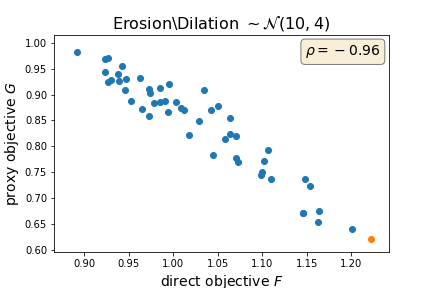}
\includegraphics[width=0.49\textwidth]{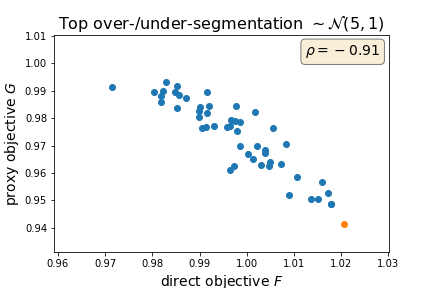}
\caption{Comparison of values of direct objective function $\boldsymbol{F}$ and the proposed proxy objective function $\boldsymbol{G}$ for two examples of mixtures of large (left plot) and small (right plot) artificial variations. The correlation coefficient is denoted as $\rho$. The point corresponding to the partition of samples in two subgroups having two different segmentation styles is highlighted with orange.}
\label{fig:scatterplots}
\end{figure}

The scatter plots showing how values of the proposed proxy objective function $\boldsymbol{G}$ are related to values of the direct objective $\boldsymbol{F}$ (experiment is described in Section \ref{subsec:obj_analysis}) are shown in Figure \ref{fig:scatterplots}. The correlation coefficient $\rho$ shows a strong dependency between the functions. Importantly, the point corresponding to $S$ is the maximum of $\boldsymbol{F}$ and the minimum of $\boldsymbol{G}$ (among the sampled points). Such results support the proposed idea of minimizing $\boldsymbol{G}$ instead of maximizing  $\boldsymbol{F}$.

The quantitative results of the proposed segmentation approach are provided in Table \ref{tab:results}. Our algorithm in all cases produces a partition of scans in subgroups which indeed have different variations. In most cases, it produces such a grouping, that all scans in a subgroup have the same segmentation style. Consequently, a better performance (in terms of both DSC and SDSC) is achieved when training separately on the two found subgroups of scans compared to training on all of them in all conducted experiments. This improvement is larger for the experiments with variations of a larger magnitude. Also, the improvement is larger for the experiments with global variations (erosion/dilation and global shifts), than for local ones, as the global variations transform the ground truth masks by a larger number of pixels. In the experiment with three different variations, the number of misclassified patients is larger and, consequently, the improvement is smaller compared to the experiment with only two of these variations (over-under-segmentation of the top part $\sim\mathcal{N}(10,4)$).

The qualitative results are demonstrated in Fig.~\ref{fig:samples}. The images show that the contour produced by a network trained on a specific style of segmentation is closer to the ground truth contour than the one produced by a network trained on a mixture of styles. Moreover, it demonstrates that a neural network can learn large variations, not necessarily aligned with the organ anatomy if such a variation is consistently present in the training data. For variations of a larger magnitude, the difference between contours produced by a network trained on mixture of segmentation styles and contours produced by networks trained on scans having a particular segmentation style is more substantial. This is demonstrated by quantitative results (in Table \ref{tab:results}) as well: the improvement gained by training networks separately is larger for the experiments with variations of a larger magnitude.

We found, that in our experiments CS-GOMEA requires 250 evaluations to converge: 200 of them are performed in the warm-up stage, 50 are used for the real search process. Function evaluations in the warm-up stage are performed in parallel. One function evaluation takes $\approx3$ minutes (we use four NVIDIA RTX 2080 Ti cards for training the neural networks). The total computational time of one experiment is $\approx$ 5 hours. 

\vspace{1cm}

\setlength\tabcolsep{2.5pt}
\begin{table}[h]
\centering
\begin{tabular}[width=\textwidth]{|c|l|c|c|c|c|c|c|c|}
\hline
\multicolumn{2}{|c|}{\textbf{Variations}} & & \multicolumn{2}{c|}{\textbf{Mixture}} & \multicolumn{2}{c|}{\textbf{Specific styles}} & \multicolumn{2}{c|}{\textbf{Improvement (\%)}} \\ \hline
     \textbf{Operation} & \textbf{Size(px)}     & \textbf{Misclass.}          &  \textbf{DSC}         &  \textbf{SDSC}         &  \textbf{DSC}         & \textbf{SDSC}          & \textbf{DSC}           & \textbf{SDSC}         \\ \hline
           erosion/dilation  & $\sim\mathcal{N}(10,4)$&     0      &    0.65      &      0.67     &   0.79         &    0.81       &    \textbf{22.91}      &   \textbf{20.49}        \\ \hline
           erosion/dilation  & $\sim\mathcal{N}(5,1)$&     0      &    0.78       &  0.79           & 0.83           &    0.84       &  \textbf{6.36}          &    \textbf{6.64}       \\ \hline
           up/down shift & $\sim\mathcal{N}(10,4)$&      0     &    0.70       &      0.72     &    0.82       &      0.84     &       \textbf{17.43}    &     \textbf{17.69}      \\ \hline
           up/down shift & $\sim\mathcal{N}(5,1)$ &      0     &     0.80      &     0.81      &      0.84     &    0.86       &       \textbf{6.06}   &     \textbf{6.26}      \\ \hline
           bottom over-/under-segmentation & $\sim\mathcal{N}(10,4)$ &     0      &    0.76       &  0.78         &   0.83        &        0.85   &      \textbf{8.2}     &     \textbf{8.16}      \\ \hline
           bottom over-/under-segmentation & $\sim\mathcal{N}(5,1)$ &     1      &    0.82       &  0.83         &   0.83        &        0.85   &      \textbf{2.16}     &     \textbf{1.90}      \\ \hline
           top over-/under-segmentation & $\sim\mathcal{N}(10,4)$&      0     &  0.78         &     0.81      &     0.83      &       0.84    &     \textbf{5.83}      &    \textbf{4.16}       \\ \hline
           top over-/under-segmentation & $\sim\mathcal{N}(5,1)$&      3     &  0.83         &     0.85      &     0.84      &       0.86    &     \textbf{1.27}      &    \textbf{1.27}       \\ \hline
          \shortstack{top over-/top under-\\ /bottom under-segmentation} & $\sim\mathcal{N}(10,4)$ &     3      &    0.76       &        0.78   &      0.79     &    0.81       &      \textbf{3.57}     &     \textbf{3.60}      \\ \hline

\end{tabular}
\caption {Main results. In the misclass. column the numbers of scans which were put in the wrong subgroup are given, based on how segmentation variations were created. DSC and SDSC scores are rounded to two decimal places.}

\label{tab:results}
\end{table}
\def\imgsize{4.1cm}
\begin{figure}[h]
\centering
\begin{subfigure}{.25\textwidth}
    \centering
    \captionsetup{justification=centering,margin=0.3cm}
    \includegraphics[height=\imgsize]{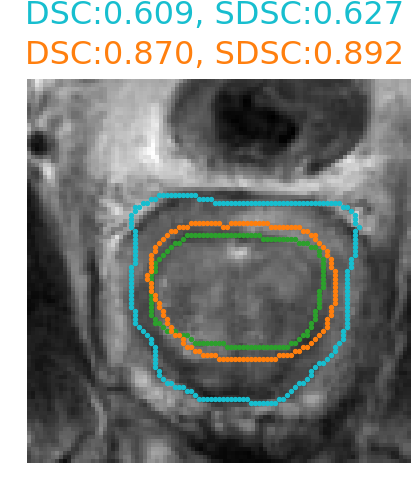}
    \caption[short]{Erosion/dilation, 10}
\end{subfigure}%
\begin{subfigure}{.25\textwidth}
    \centering
    \captionsetup{justification=centering,margin=0.3cm}
    \includegraphics[height=\imgsize]{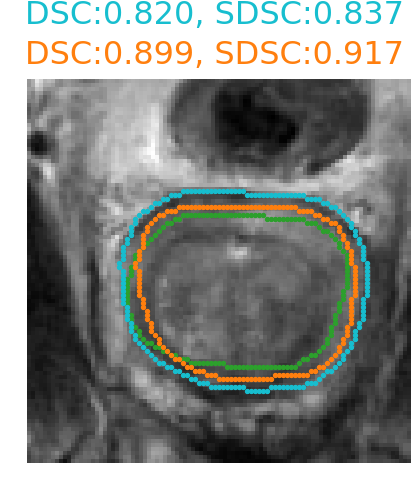}
    \caption[short]{Erosion/dilation, 5}
\end{subfigure}%
\begin{subfigure}{.25\textwidth}
    \centering
    \captionsetup{justification=centering,margin=0.3cm}
    \includegraphics[height=\imgsize]{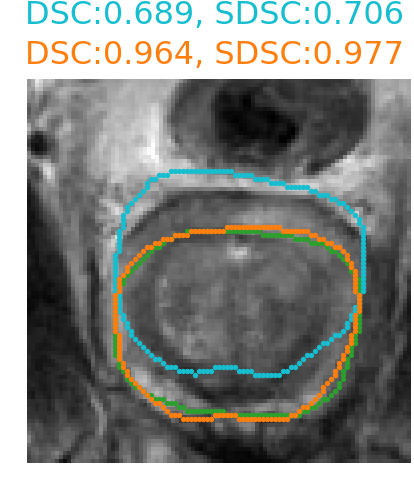}
    \caption[short]{Up/down shift, 10}
\end{subfigure}%
\begin{subfigure}{.25\textwidth}
    \centering
    \captionsetup{justification=centering,margin=0.3cm}
    \includegraphics[height=\imgsize]{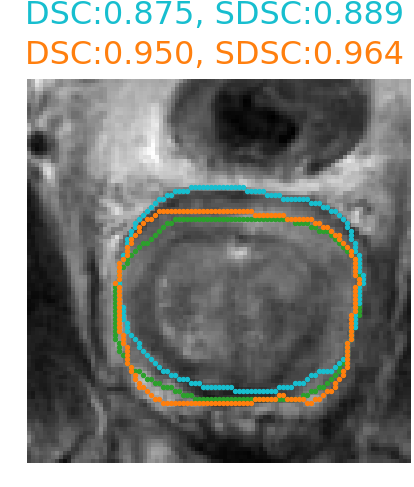}
    \caption[short]{Up/down shift, 5}
\end{subfigure}%
\vspace{0.5cm}
\begin{subfigure}{.25\textwidth}
    \centering
    \captionsetup{justification=centering,margin=0.12cm}
    \includegraphics[height=\imgsize]{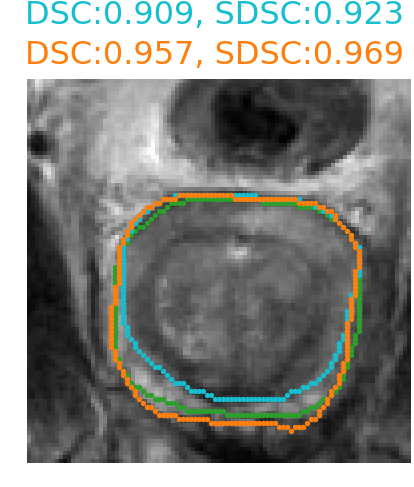}
    \caption[short]{Bottom over-/under-\\segmentation, 10}
\end{subfigure}%
\begin{subfigure}{.25\textwidth}
    \centering
    \captionsetup{justification=centering,margin=0.1cm}
    \includegraphics[height=\imgsize]{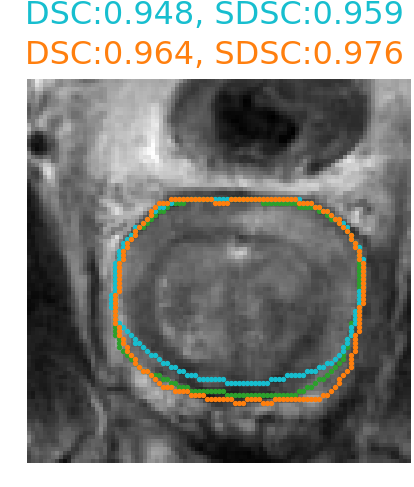}
    \caption[short]{Bottom over-/under-\\segmentation, 5}
\end{subfigure}%
\begin{subfigure}{.25\textwidth}
    \centering
    \captionsetup{justification=centering,margin=0.3cm}
    \includegraphics[height=\imgsize]{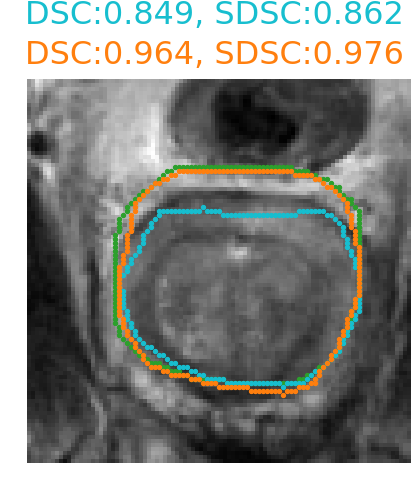}
    \caption[short]{Top over-/under-\\segmentation, 10}
\end{subfigure}%
\begin{subfigure}{.25\textwidth}
    \centering
    \captionsetup{justification=centering,margin=0.3cm}
    \includegraphics[height=\imgsize]{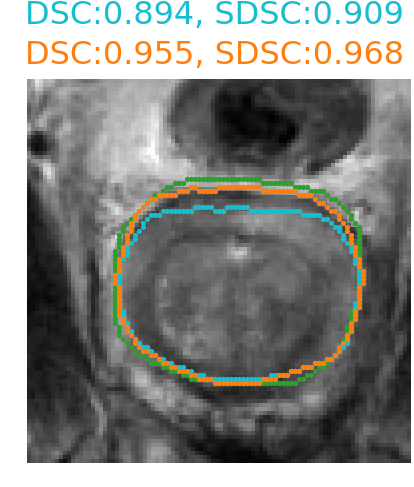}
    \caption[short]{Top over-/under-\\segmentation, 5}
\end{subfigure}%
\vspace{0.5cm}
\begin{subfigure}{\textwidth}
    \centering
    \includegraphics[width=3cm]{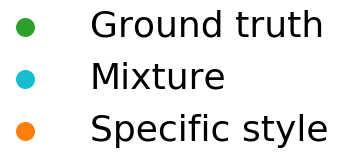}
\end{subfigure}%

\caption[short]{Samples of a slice (zoomed-in) from experiments with different variations. The DSC and SDSC scores in the first and second rows above each image correspond to scores in case of training on the mixture of styles and training on the subgroups of scans obtained by the optimization algorithm respectively. The ground truth contour (illustrated in green) represents a simulated style of segmentation: the initial ground truth contour after applying a transformation.}
\label{fig:samples}
\end{figure}

\section{Discussion}
We propose a novel approach for automatic medical image segmentation with deep learning algorithms: instead of training one universal model for all available data we propose to train separate networks for each segmentation style. Such a technique has the potential to provide better segmentation quality and better acceptance of automatic methods in clinical practice as it produces a set of predicted contours, from which a doctor can choose the preferred one.

In this paper, we use a modern CE-Net neural network. However, our approach is not limited to a particular neural network architecture. Moreover, it is not limited to deep learning-based segmentation methods. Potentially, if a new state-of-the-art algorithm for automatic segmentation appears, it can be integrated into the proposed method as long as it uses a general machine learning paradigm of learning from data. We provide proof of principle for our approach for prostate segmentation data on MRI scans. Nevertheless, the same concept could be applied to other organs and medical imaging types as well. Moreover, our approach is not limited to a particular optimization algorithm. However, considering the time-consuming nature of function evaluations of the optimization problem (one evaluation takes $\approx$ 3 minutes), it is important that the used optimization algorithm is capable of finding a good solution within a limited number of function evaluations.

In our experiments we used a dataset of moderate size: it contains 32 scans. We suppose that applying our approach to larger datasets (more than 50 scans) can provide even better results as it will allow detecting more subtle variations. In case of a small dataset, the performance of training on a larger number of subgroups results in the number of scans in these subgroups to not be sufficient to accurately learn a specific segmentation style. In the experiment with three different variations, the improvement is smaller compared to the experiment with only two of these variations and it can be explained by the lack of data in the subgroups: seven scans is perhaps not enough to accurately learn a specific segmentation style.

Our algorithm shows good results on artificially generated variations. The main assumption of the approach is that there exist different consistent segmentation styles. That is a reasonable assumption, but in case the observers are inconsistent (the variations are random) the applicability of our approach is limited and needs further study. Another assumption is that the names of observers are unknown. Such situation is not uncommon in clinical practice. Nevertheless, even if for each scan the name of the observer, who made the segmentation, is known, our algorithm has an added value. Different observers may have similar styles of segmentation, while one observer may have different styles, depending on extra information about a patient. Thus, the task of revealing subgroups of scans segmented in different styles is still non-trivial and our algorithm can be applied to solve it. 

Applying the algorithm to real clinical data is left for future research. The potential of the algorithm regarding application to other sources of variation in the training data (e.g. different scanning devices) also needs to be studied in the future.

\section{Conclusions}
We have proposed a novel approach to medical image segmentation aimed at capturing observer variation in data. Specifically, to identify different segmentation styles automatically, we proposed to automatically divide scans into subgroups (each subgroup representing a specific style of segmentation) by solving an optimization problem using the surrogate-assisted genetic algorithm CS-GOMEA.
In experiments with artificial segmentation variations added to the data (representing multiple distinguishable segmentation styles), our approach was found to be able to detect the subgroups automatically and build multiple segmentation models pertaining to these styles resulting in a substantial performance increase of auto-segmentation. Our approach is designed to find the best possible split of data to improve overall segmentation quality. Thus, it can be applied to any dataset and will likely produce such subgroups that training on them separately provides better results. Moreover, the proposed approach can be used with any segmentation method and our approach produces a grouping that leads to better results. Finally, we note, that our approach can be extended to also consider other sources of variation in data (e.g., different scanning devices).
\clearpage

\section{Acknowledgments}
This work is part of the research programme Commit2Data with project number 628.011.012, which is financed by the Dutch Research Council (NWO).

\bibliography{main} 
\bibliographystyle{spiebib} 

\end{document}